\documentclass[10pt,twocolumn,letterpaper]{article}

\usepackage{iccv}              % To produce the CAMERA-READY version
\usepackage{multirow, makecell}
\usepackage{tablefootnote}
\usepackage{multicol}
\usepackage{vwcol}

\usepackage{amsthm}

\usepackage[dvipsnames]{xcolor}
\usepackage{soul}
\usepackage{mdframed}

\newcommand{\supploc}[1]{Supp.\ Sec.\ #1}
\newcommand{\modified}[1]{#1}

\newcommand{\notebox}[1]{ }
\newcommand{\noteboxdone}[1]{ }

\newcommand{\hide}[1]{}
\colorlet{second_color}{yellow!30}

\DeclareMathOperator{\DS}{DS}

\DeclareMathOperator{\DConv}{DConv}

\newcommand{\hm}[1]{#1}  % highlight modified
\definecolor{iccvblue}{rgb}{0.21,0.49,0.74}

\usepackage[pagebackref,breaklinks,colorlinks,allcolors=iccvblue]{hyperref}
\def\paperID{10427} % *** Enter the Paper ID here
\def\confName{ICCV}
\def\confYear{2025}

\title{Towards a Universal Image Degradation Model via Content-Degradation Disentanglement}

\author{
  Wenbo Yang \\
University of Waterloo \\
{\tt\small w243yang@uwaterloo.ca}
\and
  Zhongling Wang\thanks{Work done at the University of Waterloo.} \\
AI Center-Toronto, Samsung Electronics \\
{\tt\small z.wang2@samsung.com}
\and
  Zhou Wang\\
University of Waterloo \\
{\tt\small zhou.wang@uwaterloo.ca}
}

\begin{document}
\maketitle
\begin{abstract}
  %Image degradation synthesis is essential for applications ranging from image restoration to simulating artistic effects, like film grain. Existing degradation models lack generalizability because they are designed to generate one specific or narrow set of degradations, which often require user-provided degradation parameters. Consequently, they cannot synthesize degradations beyond their initial design or adapt to other applications. To address this, we propose the \textbf{first} universal degradation model that can synthesize a broad spectrum of complex and realistic degradations that contain both homogeneous (global) and inhomogeneous (spatially varying) components. Our model automatically extracts and disentangles homogeneous and inhomogeneous degradation features, which are later used for degradation synthesis without user intervention. To separate degradation information from images, we propose a disentangle-by-compression method. To model inhomogeneous components presented in complex degradations, we proposed two novel modules for extracting and incorporating inhomogeneous degradation. We demonstrate the model’s accuracy and adaptability in film-grain simulation and blind image restoration tasks.\footnote{The code and dataset of this project will be released.}

Image degradation synthesis is highly desirable in a wide variety of applications ranging from image restoration to simulating artistic effects. Existing models are designed to generate one specific or a narrow set of degradations, which often require user-provided degradation parameters. As a result, they lack the generalizability to synthesize degradations beyond their initial design or adapt to other applications. Here we propose the \textbf{first} universal degradation model that can synthesize a broad spectrum of complex and realistic degradations containing both homogeneous (global) and inhomogeneous (spatially varying) components. Our model automatically extracts and disentangles homogeneous and inhomogeneous degradation features, which are later used for degradation synthesis without user intervention. A disentangle-by-compression method is proposed to separate degradation information from images. Two novel modules for extracting and incorporating inhomogeneous degradations are created to model inhomogeneous components in complex degradations. We demonstrate the model’s accuracy and adaptability in film-grain simulation and blind image restoration tasks.  The demo video, code, and dataset of this project will be released at \href{https://github.com/yangwenbo99/content-degradation-disentanglement}{Yang's GitHub account} .  % \footnote{The demo video, code and dataset of this project will be released.  An anonymized version of the demo video is submitted as supplementary material.}

\end{abstract}
    
\section{Introduction}
\label{sec:intro}

\begin{figure}[t]
  \centering
  \includegraphics[width=0.99\columnwidth]{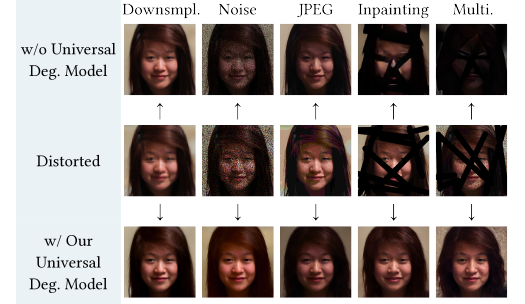}
  \caption{Demonstration of the applicability of our universal image
    degradation model in GAN inversion-based image restoration.  When a
    fixed degradation model is used, only the modeled degradation
    (downsampling, first column) receives high-quality results.  However,
    when our universal degradation model is used, all results are of
    pleasant quality.  Best to be viewed zoomed in. 
  }
  \label{fig:cover}
\end{figure}

\begin{figure*}[t]
  \centering
  \includegraphics[width=0.90\textwidth]{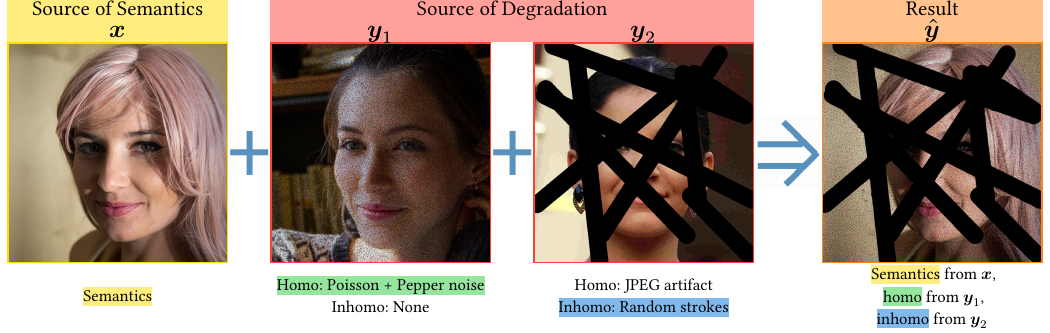}
  \caption{% \hl{first two images have similar global distortion, it's hard to say which one is the source of global. Also include the pristine image.}
    Visualization of our models' degradation disentanglement and transfer ability. 
    Our model can extract homogeneous and inhomogeneous degradation
    information from the same or different distorted image(s) and
    reapply it to a pristine image.
    % When IDA or the IDE network is removed from the model, its ability
    % of reproducing degradation becomes inhibited.  
  }
  \label{fig:abl_2}
\end{figure*}

% What is image degradation (w. example in each stage)?
% Images in real life are subject to a wide range of degradations that impact their perceptual qualities during acquisition, processing, and communication. These include focus and motion blur from improper capture setups \cite{spaq}, inevitable photon shot noise, and other noise sources introduced by camera signal processing \cite{noise_low_light}.  Moreover, during communication, repeated downsampling and compression can introduce complex artifacts \cite{deepdebanding2022}. Real-world images often contain complex combinations of these degradations.
% Why do we need to reproduce these degradations? Shouldn’t pristine images be more desirable?
%Image degradation synthesis has a broad range of applications. For example, some image restoration approaches use it indirectly by first generating pristine-distorted image pairs through degradation models, which are then employed to train restoration models in a supervised manner \cite{Zou_2022_CVPR,Zhang_2021_ICCV_2,Ji_2020_CVPR_Workshops,sirr_wang}.
%Other restoration models directly incorporate degradation synthesis as a module within their frameworks \cite{7780421,dehazenet,film_grain_0,robust_stylegan,danet,noise_flow}. Additionally, accurate degradation synthesis is valuable for artistic effects creation. For example, 
%movie producers firmly believe
%precise film grain synthesis enhances the viewer experience \cite{av1_filmgrain}. Efficient haze and rain synthesis can replace the need for a complex physical engine in computer graphics.

Image degradation synthesis plays an important role in numerous applications across multiple domains. In image restoration, it is often used indirectly to generate pristine-distorted training pairs for supervised learning
\cite{Zou_2022_CVPR,Zhang_2021_ICCV_2,Ji_2020_CVPR_Workshops,sirr_wang}, while some approaches incorporate it directly as an integral module within restoration frameworks
\cite{7780421,dehazenet,film_grain_0,robust_stylegan,danet,noise_flow}. Beyond restoration, accurate degradation synthesis also plays a significant role in artistic applications. For example, cinematographers rely on precise film grain synthesis tor enhance viewer experience \cite{av1_filmgrain}, and efficient haze and rain synthesis offers a computationally efficient alternative to complex physical simulations in computer graphics.

% What's wrong with existing models
However, existing degradation models, whether handcrafted, statistical, or deep learning-based, are predominantly designed for specific degradation types or narrow degradation sets, as summarized in \Cref{tab:models}.
These models often employ restrictive design choices to optimize performance
for particular degradations, sacrificing generalizability across diverse scenarios.
Consequently, they \modified{cannot be adapted to} synthesize degradations beyond
their original design. 
However, in many downstream applications, the ability to synthesize
multiple types of degradation is essential.  A set of visual examples in
image restoration is given in \cref{fig:cover}. 
Furthermore, most existing models necessitate user-specified degradation parameters,
limiting their applicability in contexts where \modified{frequent user input is impractical},
such as film grain transfer or blind image restoration.
While some degradation models can estimate
the degradation parameters, they only apply to simple, well-defined
degradations, failing to extend to complex real-world degradations that
are usually hard to describe parametrically. Furthermore, due to the
intrinsic design limitations, current methods 
cannot synthesize
\textit{inhomogeneous} degradations that vary spatially, a common characteristic
of real-world distortions (examples are available in \supploc{7}). 
Among existing works, only \citet{bovik_2020} attempted to
develop a model without degradation-specific design. However, their
approach still requires training separate models for each distortion due
to the lack of degradation representation and cannot synthesize
inhomogeneous degradations.

Therefore, it is desirable to design a \textit{universal} degradation model that can generalize to \textit{complex} (combinations of) degradations, both homogeneous and \textit{inhomogeneous}. However, achieving this is challenging for two main reasons. First, complex degradations are harder to separate from degraded images than simpler ones.  Simple degradation models often assume linear (or affine) and content-independent degradation processes. However, real-world complex degradations are typically nonlinear and content-dependent, complicating the extraction of pure degradation information free of image content. Second, extracting and synthesizing inhomogeneous degradations presents additional challenges, as these spatially varying effects are more difficult to model than homogeneous ones.

In this paper, we present the first \textit{universal image degradation
model} capable of encoding and synthesizing complex combinations of homogeneous and inhomogeneous image degradations. 
As demonstrated in  \Cref{fig:abl_2}, our model effectively separate
degradation information from distorted images' contents, disentangle inhomogeneous and homogeneous degradations, and transfer degradations to other pristine images. To encourage the separation of degradation information from image content, we propose a novel \textit{disentangle-by-compression} method to regularize the model behavior. 
To handle inhomogeneous degradations, we further propose two network modules, namely Inhomogeneous Degradation Embedding Network (IDEN)  and Inhomogeneous Degradation Aware (IDA) network layer, which empowers the encoding and decoding of inhomogeneous degradations, respectively.
\modified{Our model advances explainable degradation modeling and enables detailed analysis of degradation components. It also finds applications in a wide range of image restoration tasks and in the visual content generation and distribution industry.}
%\modified{Our model has practical applications in the film industry and image restoration, while also advancing the understanding of degradation processes and enabling detailed analysis of degradation components.}

We demonstrate our model's effectiveness
through extensive ablation studies and in downstream applications such
as film grain simulation and blind image restoration. For the latter,
our model can be used as a drop-in replacement for the degradation
synthesis module in \modified{generative model inversion-based}
image restoration methods \modified{for any type of image contents (as
long as the inverted generative model supports)}, converting existing non-blind image restoration models into blind ones. Blind models do not require any degradation information from the user and have broader use cases in real-world applications, especially in cases where the degradation is complex and hard to describe parametrically. 
In summary, our key contributions are:
\begin{itemize}
  \item We propose the first universal degradation model capable of
    handling complex combinations of homogeneous and inhomogeneous
    degradations.
  \item We introduce a novel disentangle-by-compression method that
    separates degradation information from distorted images' contents and encourages the components in degradation embedding to be independent.
  \item When used as a drop-in module for inversion-based image
    restoration, it enables, for the first time, generative model
    inversion in a blind setting to handle unknown and complex
    distortions.% \modified{for any contents the inverted generative model supports}.
\end{itemize}

\section{Related Works}
\label{sec:lr}

% \subsection{Real-world degradation simulation}
% \label{sec:deg_sim}

\begin{table}[t]%htbp]
\centering
% \sffamily
\footnotesize % Make the font size smaller
%\begin{tabular}{|p{1.6in}|p{0.7in}|p{0.5in}|}
\begin{tabular}{|p{0.35in}|p{0.65in}|p{1.2in}|l|}
\hline
\multicolumn{2}{|l|}{\textbf{Degradation Type}} & \textbf{Model} & \textbf{Mdl Type}                                                                                                          \\ \hline
\parbox[t]{2mm}{\multirow{10}{*}{\rotatebox[origin=c]{90}{Homogeneous}}}
                                                & \multirow{3}*{Noise}         & \cite{Zou_2022_CVPR},\cite{noise_low_light},\cite{Zhang_2021_ICCV_2},\cite{pg_denoising},\cite{pg_modeling},\cite{7780421} & {Stat-based }                   \\ \cline{3-4}
                                                &                              & CA-NoiseGAN \cite{Chang_2020},Noise Flow \cite{noise_flow},DANet \cite{danet}                                              & \multirow{2}*{Learned }         \\ \cline{2-4}
                                                & Downsample + Noise           & RealSR \cite{Ji_2020_CVPR_Workshops}                                                                                       & \multirow{2}*{Learned }         \\ \cline{2-4}
                                                & Rain                         & \cite{sirr_wang}                                                                                                           & Learned                         \\ \cline{2-4}
                                                & Old Photo                    & \cite{rephotography}                                                                                                       & Stat-based                      \\ \cline{2-4}
                                                & \multirow{2}*{Grain }        & \cite{film_grain_0}                                                                                                        & Stat-based                      \\ \cline{3-4}
                                                &                              & \cite{film_grain_1},\cite{film_grain_2}                                                                                    & Learned                         \\ \cline{2-4}
                                                & Various single               & \cite{bovik_2020}                                                                                                          & Learned                         \\ \cline{1-4}
\multirow{3}*{Inhomo-}                                          & \multirow{3}*{Haze}          & \cite{dark_channel}, \cite{max_contrast}, \cite{color_atten}, \cite{hue_disparity}                                         & Stat-based                      \\ \cline{3-4}
                                                &                              & DehazeNet \cite{dehazenet}, AOD-Net \cite{aod-net}                                                                         & \multirow{1}*{Hybrid }          \\ \cline{1-4}
Both                                            & \textbf{Complex combination} & \multirow{2}*{\textbf{Ours}}                                                                                               & \multirow{2}*{\textbf{Learned}} \\ \hline
\end{tabular}
\caption{
  A review of representative degradation models.  Most of the models focus on single homogeneous distortion.
  % Models marked with ``*'' requires multiple training examples with identical degradation settings, and those with ``**'' has other constraints on training examples. 
}
\label{tab:models}
\end{table}

\Cref{tab:models} summarizes representative degradation models. Most existing models focus on a single distortion type, while some handle basic fixed combinations of distortions. 
For instance, some super-resolution models \cite{Zhang_2021_ICCV,real-esrgan} considers common degradations in the imaging pipeline such as blur, downsampling, noise and compression artifacts. However, each degradation component is modeled independently. 
Some components are handcrafted, requiring strong domain knowledge specific to the downstream task (super-resolution in this case). 
These degradation models cannot be easily transferred to other degradation types. 

Attempting to model all possible degradations individually is impractical, highlighting the need for a universal degradation model that can generalize across degradations without requiring domain-specific knowledge from downstream applications.  However, constructing such a model is challenging: complex combinations of distortions are typically intractable for explicit modeling and have not yet been effectively simulated through machine learning-based approaches. To our knowledge, only \cite{bovik_2020} attempted to construct a non-distortion-specific image degradation model. However, their approach still requires training separate models for each distortion type, as it does not accept degradation information as input. Furthermore, they did not consider the stochastic effects of image distortions, particularly for image noise, resulting in similar generated noise patterns across images.

\hide{
  BSRGAN \cite{Zhang_2021_ICCV} presented an alternative approach to the
  intractability of the difficulty in accurately recreating complex
  degradation for training image restoration model
  by creating an over-comprehensive set of degradations
  that is expected to contain most degradations presented in the testing
  set, but this requires strong domain knowledge and may not transfer
  easily to other applications.
}

\hide{
BSRGAN \cite{Zhang_2021_ICCV} presented an alternative approach to the
intractability of the difficulty in accurately recreating complex degradation for training image restoration model
by creating an over-comprehensive set of degradations that is expected
to contain most degradations presented in the testing set.  It augments
each of its training images with 9 to 12 randomly shuffled distortion. 
However, generating the list of degradations requires very strong domain
knowledge and the list may not be easily transferred to other
applications. 
}

Another major limitation of most existing degradation models is their assumption that degradations are homogeneous across the entire image. This simplification often fails to capture the complexity of real-world scenarios, where degradations can vary significantly across different regions of an image. Some haze simulation methods attempt to address this issue by using a depth map to model inhomogeneous haze. However, this approach is specific to haze and cannot be generalized to other types of degradations.

\begin{figure}[t]
  \centering
  \includegraphics[width=1.0\columnwidth]{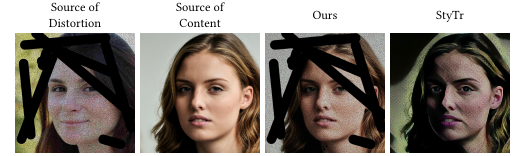}
  \caption{
    Visual results for StyTr2 \cite{stytr2} retrained on the distortion
    dataset. It fails at separating distortion from content, hence,
    alters skin tone and texture while not transferring distortions
    (more examples in \supploc{3.5}).
  }
  \label{fig:st}
\end{figure}

% \subsection{Style transfer}
Another related area is Style Transfer (ST), which aims to impose the artistic
style of a source image to the content of a target image. 
\modified{However, most ST methods require training a separate model for
each specific artistic style. A few SOTA methods handle multiple
styles by extracting style information from a separate input image
\cite{style_transfer_review,CAI2023108723,stytr2}.  
Nevertheless, image degradations exhibit distinct characteristics from artistic styles, necessitating specialized architecture for effective modeling. As a result, SOTA ST models (e.g., StyTr2 \cite{stytr2}) exhibit significant limitations (\cref{fig:st}): they create undesirable changes to image content (e.g., skin tone), are poor at reproducing realistic noise patterns, are incapable of segregating homogeneous and inhomogeneous degradations, and fail at handling inhomogeneous degradations due to architectural limitations.}

%However, similar to the limitations of current degradation models, most style transfer methods require training a separate model for each specific artistic style, \modified{while only a few SOTA methods, such as StyTr2 \cite{stytr2}, can handle multiple styles by extracting style information from a separate input image \cite{style_transfer_review,CAI2023108723}.  Nevertheless, image degradations exhibit distinct characteristics from artistic styles, necessitating specialized architecture for effective modeling. Even SOTA ST models like StyTr2 \cite{stytr2} exhibit significant limitations (\cref{fig:st}): they significantly alter image content (e.g., skin tone), cannot reproduce realistic noise patterns, and fail on inhomogeneous degradations due to architectural limitations. Furthermore, they also cannot segregate homogeneous and inhomogeneous degradations.}

\section{Proposed Method}
\begin{figure}[t]
  % Let it flow to save some space
  \centering
  \includegraphics[width=0.95\columnwidth]{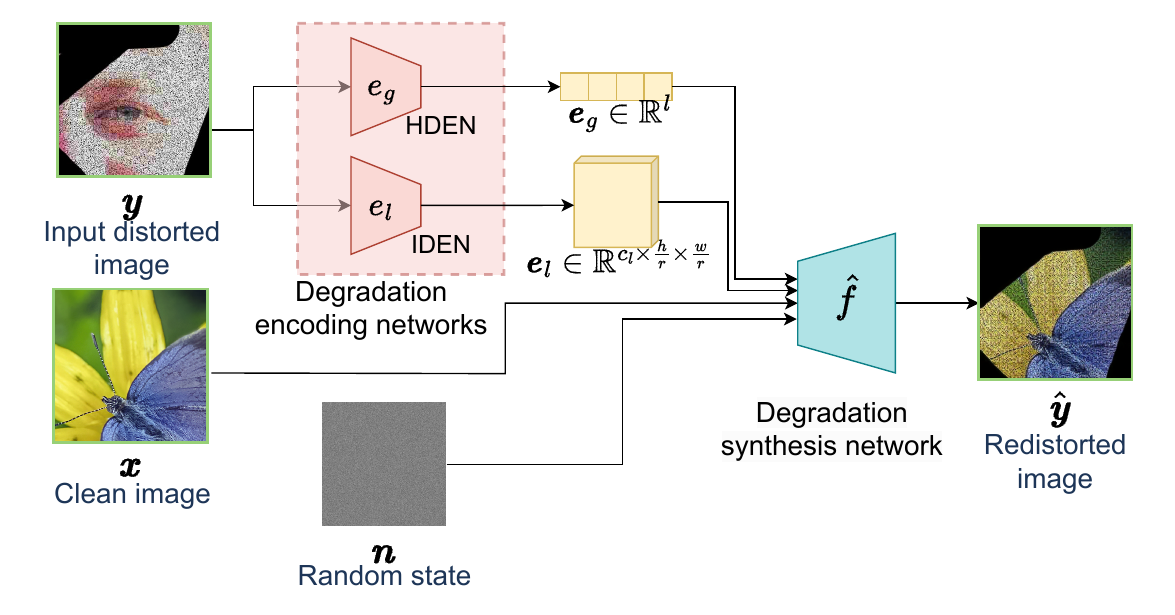}
  \caption{
    The overall architecture of our degradation encoding and decoding
    network (during the test phase).  
    % The degradation encoders \({e}_g\) and
    % \({e}_l\) accept a distorted image and generate
    % corresponding degradation embeddings. These embeddings
    % are combined with a clean image to generate another distorted image with
    % reproduced degradations.  
  }
  \label{fig:base}
\end{figure}

We note that all distorted process can be described by
the following formulation:
\begin{equation}
\mathbf{y} = f\left( \mathbf{x}, \mathbf{d}, \mathbf{n} \right),
\end{equation}
where \(\mathbf{x}\) is a pristine image, 
\(\mathbf{d}\) is
the degradation information,
\(\mathbf{n}\) is a random state, 
and \(f\) is an operator synthesizing distortions based on the
information.  
Due to the complexity of real-life degradations, it is intractable to
explicitly construct a statistical model.
Hence, end-to-end learned degradation encoding and decoding
Convolutional Neural Networks (CNNs) are necessary. 
\Cref{fig:base} shows an overview of our framework. 
A network \(\hat f\) is trained to simulate \(f\), which degrades a
pristine image \(\mathbf{x}\) with the degradation extracted by
a homogeneous and inhomogeneous degradation encoding networks \(e_g\) and \(e_l\), respectively:
\begin{equation}
  \hat{\mathbf{y}} := \hat f\left( 
    \mathbf{x}, \mathbf{e}_g, \mathbf{e}_l, \mathbf{n} 
  \right),
\end{equation}
where \(\mathbf{e}_g := e_g(\mathbf{y})\),
and \(\mathbf{e}_l := e_l(\mathbf{y})\).  
\modified{%
  With our novel training strategy,
  our model learns to transfer degradation from one image
  \(\mathbf{y}^{(1)}\) (or images \( \mathbf{y}^{(1)} \) and
  \(\mathbf{y}^{(2)}\))
  to another image \(\mathbf{x}^{(0)}\) without 
  being supervised  on the triplet \(\left( \mathbf{y}^{(1)},
    \mathbf{x}^{(0)}, \mathbf{y}^{(0)} \right)\) (or
    \(\left( \mathbf{y}^{(1)}, \mathbf{y}^{(2)}, \mathbf{x}^{(0)},
      \mathbf{y}^{(0)} \right)\)).
}%
Detailed explanations of \(e_g, e_l, \hat f\) and training strategies are provided in the following sections.

\subsection{Homo-/inhomogeneous degradation encoding}

While most works ignore the spatial variance of image degradations, these inhomogeneous degradations occupy a large portion of real-world degradations.
Hence, our model employs two degradation encoding networks: the \textit{homogeneous degradation embedding network} (HDEN) \(e_g\) and the \textit{inhomogeneous degradation embedding network} (IDEN) \({e}_l\).
The homogeneous degradation embedding (HDE) \(\mathbf{e}_g\) captures spatial-invariant degradation information, and the inhomogeneous degradation embedding (IDE) \(\mathbf{e}_l\)  retains spatial structure for local degradations.

\begin{figure}[t]
  \centering
  \includegraphics[width=0.98\columnwidth]{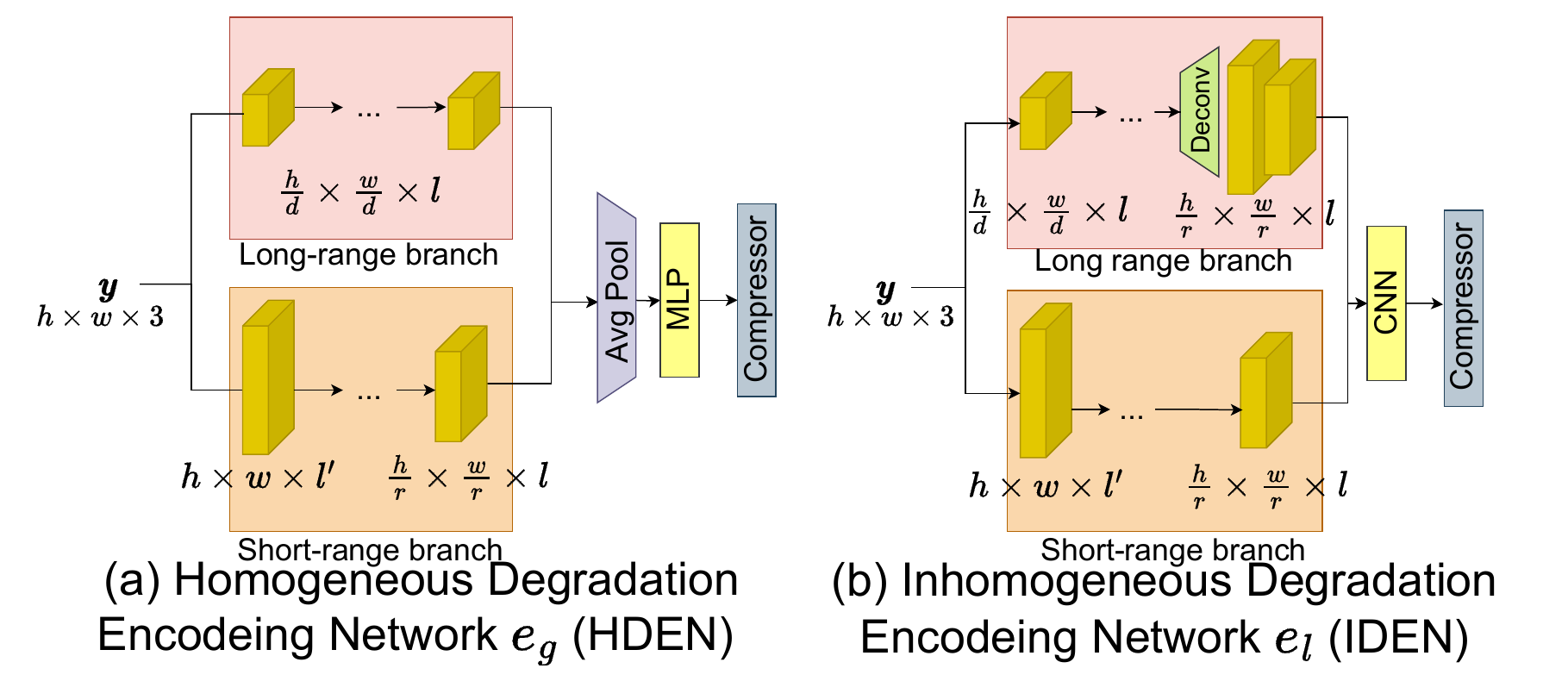}
  \caption{
    The architecture of the two Degradation Encoding Networks: HDEN and IDEN.
  }
  \label{fig:des}
\end{figure}

As illustrated in \cref{fig:des}, the HDEN has a dual-branch architecture. To accommodate degradations requiring different receptive fields (e.g., blur vs. noise), we incorporate (a) a short-range branch operating at the input image scale and (b) a long-range branch operating on a downsampled (by a factor \(d\)) resolution. 
This design, inspired by \citet{zhou2022}'s encoder, can effectively capture globally consistent degradation's characteristics.  
% However, due to the missing spatial structure in the result, HDEN struggles to capture inhomogeneous degradations.   
Additionally, to capture inhomogeneous degradations, we introduce IDEN, which has a modified long-range branch and a tail module that retains spatial structure. 
% IDEN upsamples in the long-range branch to match the spatial resolution
% of the short-range branch.  
The post-processing Multi-layer perception
(MLP) in IDEN is replaced with a small CNN to efficiently combine long-range and short-range information on the spatial domain.

\subsection{Homo-/inhomogeneous degradation synthesis}

\hide{
  Key points
  \begin{enumerate}
    \item Decompression module is part of the ``network''
    \item VUNet
    \item Usage of random state
    \item Modifications: Conv. for spatial dependency
  \end{enumerate}
}

\begin{figure*}[tb]
  \centering
  \includegraphics[width=0.95\textwidth]{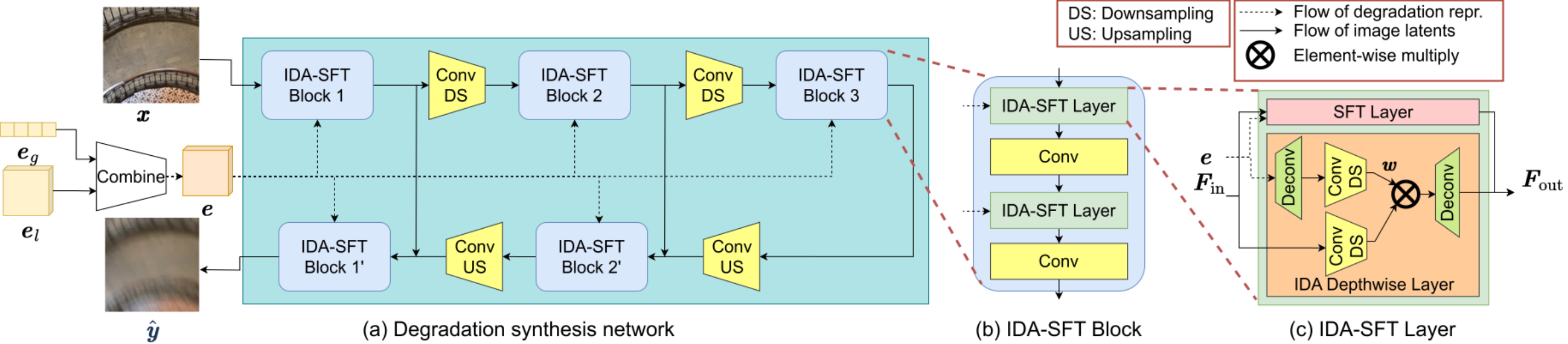}
  \caption{
    The degradation synthesis network. 
  }
  \label{fig:vunet}
\end{figure*}

Our degradation synthesis network has a U-Net structure \cite{noise_transfer} consisting of 
several IDA-SFT blocks, as shown in
\Cref{fig:vunet}. 
The global degradation embedding
\(\mathbf{e}_g\) 
and the local degradation embedding \(\mathbf{e}_l\)
are first combined into an intermediate degradation map
\(\mathbf{e}\), which then conditions each IDA-SFT block to degrade the input image \(\mathbf{x}\).

\hm{We identify providing a smooth channel of incorporating local
  degradation as a practical requirement for disentangling local
  degradation information from global. 
  Although performing convolution using a spatially-varying kernel is an ideal
  way of providing such channel, 
  its space complexity is prohibitively expensive. 
}%
% To more efficiently capture locally dependent degradations, 
% especially those fast-changing ones,
% it would be desirable to perform convolution using
% spatially-varying kernels. 
The main cost is attributed to the inter-channel
dependencies, while spatial dependencies are mainly handled by
depthwise components. 
To reduce the computation cost, 
we propose an inhomogeneous degradation-aware (IDA) layer: 
% in parallel to the MLPs: 
\begin{equation}
  \text{IDA}(\mathbf{F}_{\text{in}}, \mathbf{e}) = 
  \DConv\left(
  \DS(\DConv(\mathbf{e})) \odot \DS(\mathbf{F}_{\text{in}})
  \right),
\end{equation}
where
% \(\mathbf{e}'\) is the intermediate embedding \(\mathbf{e}\)
% interpolated to an appropriate scale, 
\(\DS\) is a downsampling operation achieved by convolution
with stride and \(\DConv\) is the deconvolution operation \cite{deconv} that can upsample a latent. 
% By operating on a lower spatial scale, 
We have proved in \supploc{6.1} that a single IDA layer is more expressive and flexible than the combination of four depthwise
convolution layers with spatially varying kernel. 
% \begin{proposition}
%   Given a latent vector \(\mathbf{F}_{\text{in}} \in
%   \mathbb{R}^{c \times 2h \times 2w}\). 
%   For any combination of four
%   $2\times2$ depthwise convolution layers
%   \(\left\{ C_k : k \in \mathbb{N}_4 \right\}\),  with stride 2 and 
%   spatially-varying kernel  
%   \(W^{(k)} \in \mathbb{R}^{h \times w \times c \times 2 \times 2}\),
%   suppose there exists some 
%   $u^{(k)}_{pqr}$'s such that
%   \(W^{(k)}_{ijpqr} 
%   =
%   u^{(k)}_{pqr}
%   W^{(1)}_{ijpqr}
%   \) 
%   for every \(i \in \left\{ 2, 3, 4 \right\}\). 
%   Then, 
%   there exists an \(\mathbf{e} \in \mathbb{R}^{4 c, h/2, w/2}\) such that 
%   \[
%     \left( 
%       \IDA(\mathbf{F}_{\text{in}}, \mathbf{e})
%     \right)_{p, 2 i + a, 2 j + b}
%     =
%     \left( 
%       C_{2 a + b - 2}(\mathbf{x})
%     \right)_{p, i, j},
%   \]
%   for \(a, b \in \left\{ 1, 2 \right\}\). 
% \end{proposition}
% \begin{proof}
% See \supplement.
% \end{proof}
%

The IDA layer excels at simulating
inhomogeneous degradations for our application. 
We combine it
in parallel with a Spatial Feature Transform (SFT) layer \cite{sft},
% \hm{which homogeneously fuses the feature map}
% \(\mathbf{F}_{\text{in}}\), the degradation latent (condition)
% \(\mathbf{e}\) and the random state
which is good at 
introducing random states to latents and generating homogeneous
degradations, to form the IDA-SFT layer:
% capturing high-frequency and slow-varying degradations, 
% to enhance the generation of spatially dependent distortions in
{\small
\begin{equation}
  \text{IDA-SFT}(\mathbf{F}_{\text{in}}, \mathbf{e}, \mathbf{n}) = 
  \text{IDA}(\mathbf{F}_{\text{in}}, \mathbf{e})
  +
  \underbrace{
    \alpha(\mathbf{e}) \odot \mathbf{F}_{\text{in}}
    + 
    \beta(\mathbf{e}, \mathbf{n})
  }_{\text{Spatial Feature Transform}}
  ,
\end{equation}
}
where \(\alpha\) and \(\beta\) are two small MLPs.

\subsection{Degradation disentanglement}
\label{sec:compression}

Our model achieves three key disentangling effects: separating (1)
degradation from distorted images contents, (2) inhomogeneous from
homogeneous distortions, and (3) individual degradation components.
These effects enable degradation transfer between images and direct
manipulation of degradation characteristics.
We attribute this to our
novel disentangle-by-compression approach, implemented through
\textbf{entropy regularization losses}:
\[\mathcal{L}_{\mathrm{\text{rate\_g }}} = \sum_{i}H\left( e_{g}^{(i)} \right) = - \sum_{i}\mathbb{E}\log_{2}p\left( e_{g}^{(i)} \right),\]
\[\mathcal{L}_{\mathrm{\text{rate\_l }}} = \sum_{i,j}H\left( e_{l}^{(i,j)} \right) = - \sum_{i,j}\mathbb{E}\log_{2}p\left( e_{l}^{(i,j)} \right),\]
where \(p\) denotes the probability density function. The density
% Modified here... Changing two word (is -> denote, +function).  This 
% adds two lines in the result
\(p\left( e_{g}^{(i)} \right)\) and \(p\left( e_{l}^{(i,j)} \right)\)
are estimated using \cite{balle2018} and \cite{balle2016}, respectively.
Due to the large sample size used in the estimation of \(p\), its
estimation is accurate enough (while \(p\left( \mathbf{e}_{g} \right)\)
cannot be reliably estimated, due to its high dimensionality).

We shall justify the entropy regularization loss's effects in remainder
of this section. Since \(\mathbf{e}_{l}\) and \(\mathbf{e}_{g}\) are
similar in nature, we shall drop the subscript until the discussion of
inhomogeneous and homogeneous disentanglement. The sum of entropy
\(\mathbf{e}\) can be expanded as \cite{total-correlation}
\begin{equation}
  \sum_{i}H\left( e^{(i)} \right) = \underset{\mathrm{\text{separate degrad. from img}}}{\underbrace{H\left( \mathbf{e} \right)}} + \underset{\mathrm{\text{enforce entries’ independence}}}{\underbrace{D_{\mathrm{\text{KL}}}\left( p\left( \mathbf{e} \right) \parallel q\left( \mathbf{e} \right) \right)}},
  \label{equ:entropy-a}
\end{equation}
where \(q\left( \mathbf{e} \right) := \prod_{i}p\left( e^{(i)} \right)\).
By optimizing the left-hand-side of \cref{equ:entropy-a}, we are
simultaneously minimizing the entropy of \(\mathbf{e}\) and the
KL-divergence of its distribution towards \(q\), which has the same
marginal distribution as \(p\left( \mathbf{e} \right)\) but with
independent entries. The latter makes the entries in \(\mathbf{e}\)
independent.

The following mild assumptions are made: (\textbf{A-1}) With information
about the degradation process and the clean image, the degraded image
can be exactly reconstructed, albeit the random state difference.
(\textbf{A-2}) The distribution of distortion processes is independent of the clean images'.
(\textbf{A-3}) The distortion can be inferred from the degraded image.
(\textbf{A-4}) When an appropriate \textbf{perceptual similarity loss}
\(\mathcal{L}_{\text{sim }} = d\left( \mathbf{y},\hat{\mathbf{y}} \right)\)
is used, the reconstructed distorted image \(\hat{\mathbf{y}}\) will be
be similar enough to \(\mathbf{y}\). 
(See \supploc{2.2} for detailed justifications.) 
With these assumptions, we are able
to rigorously prove (in \supploc{6.2}) that
\begin{equation}
  H\left( \mathbf{e} \right) = I\left( \mathbf{e};\mathbf{x} \right) + H\left( \mathbf{d} \right).
  \label{equ:expansion-he}
\end{equation}
Since \(H\left( \mathbf{d} \right)\) is a fixed (but unknown) constant,
by minimizing \(H\left( \mathbf{e} \right)\), the mutual information
\(I\left( \mathbf{e};\mathbf{x} \right)\)
is minimized, which encourages
\(\mathbf{e}\) to contain no information in \(\mathbf{x}\) (which
contains all content information), disentangling the degradation from
image content.

We further observe that there exists two kinds of
degradations: the homogeneous one \(\mathbf{d}_{g}\) is the same for all
geometry location, and the inhomogeneous one \(\mathbf{d}_{l}^{(i)}\) is
different for different geometry location \((i)\).
Based on our observations, we assume (\textbf{A-5}) they are
jointly independent. Then, we show in the supplementary materials that,
with a flexible enough network (and 
\(\lambda_l\) and \(\lambda_g\) satisfying mild conditions), the optimal solution for
\(\lambda_{g}\mathcal{L}_{\text{rate\_g}} + \lambda_{l}\mathcal{L}_{\text{rate\_l}}\)
guarantees that \(\mathbf{e}_{l}\) only contains spatially dependent
information, and \(\mathbf{e}_{g}\) only contains homogeneous distortion
information.

\subsection{Loss functions}

We choose DISTS \cite{dists} as our perceptual distance measure \(d\) since 
it has the least unnecessary sensitivity to noise's random state (which
cannot be perceived by human observers, details are in \supploc{6.2}). 
To ensure diverse outputs, we use SSIM \cite{ssim} as the diversity loss due to its sensitivity to random states:
\[\mathcal{L}_{\text{diver}} = - \text{SSIM}(\hat{\mathbf{y}}, \hat{\mathbf{y}}')\]
where \(\hat{\mathbf{y}}\) and \(\hat{\mathbf{y}}'\) are generated using the same inputs but different random states.
To address the limitations of general-purpose IQAs in capturing human-perceived fidelity, we incorporate an adversarial loss \(\mathcal{L}_\text{gan}\) for more realistic degradation generation.
Following conventions, our total loss function also incorporates a contrastive loss \(\mathcal{L}_{\text{contra}}\) \cite{dasr} and color loss \(\mathcal{L}_{\text{color}}\) \cite{color_loss}.
Our total loss function is 
\begin{equation}
	\begin{matrix}
		
		\mathcal{L} :=
		& 
		\mathcal{L}_{\text{sim}}
		+ 
		\lambda_g \mathcal{L}_{\text{rate\_g}}
		+ 
		\lambda_l \mathcal{L}_{\text{rate\_l}}
		+ 
		\lambda_c \mathcal{L}_{\text{contra}}
		\\ 
		&
		+ 
		\lambda_r \mathcal{L}_{\text{color}}
		+ 
		\lambda_d \mathcal{L}_{\text{diver}}
		+ 
		\lambda_g \mathcal{L}_{\text{gan}},
	\end{matrix}
\end{equation}
where all \(\lambda_\cdot\)'s are trade-off weights.

\section{Experiments}

\subsection{Degradation reproduction and transfer}

Our model is the first to encode, synthesize, and transfer arbitrary degradation combinations from existing degraded images.
\modified{It is designed to train on paired clean-distorted datasets with any type of degradation. However, the diversity of degradations and content in current datasets remains limited. Additionally, real-world datasets often lack ground-truth references, making it challenging to benchmark degradation transfer performance. To address this, we supplement real-world datasets \cite{rain_dataset, gopro} with synthetic datasets inspired by real-world scenarios for quantitative evaluations and more challenging tasks. We believe our model can serve as a baseline for degradation synthesis and transfer, encouraging the development of more comprehensive paired real-world degradation datasets.}
%\modified{It can be trained on paired clean-distorted datasets with any type of degradations. However, current datasets are limited in the diversity of degradations and contents. Moreover, real-world datasets are lack of ground-truth for benchmarking degradation transfer performance. Hence, we are limited in our evaluation by the availability of suitable datasets, despite our model's capability to handle a wide range of distortions. In addition to real-world datasets \cite{rain_dataset, gopro}, we also use real-world-inspired synthetic datasets quantitative evaluations for degradation transfer (real-world datasets) and more challenging tasks. We believe our model can act as a baseline for degradation synthesis and transfer, and incentivize the development of more paired real-world degradation datasets.}

\begin{figure}[t]
  \centering
  \includegraphics[width=0.98\columnwidth]{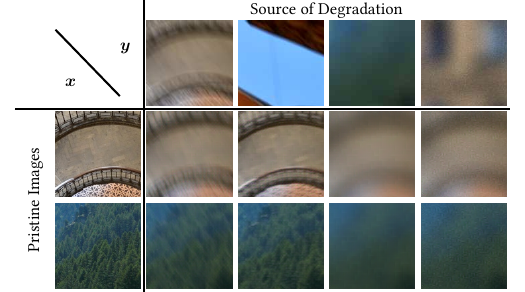}
  \caption{
    Synthetic distortion transfer. Each grid is a
    synthesized degraded image, generated by distorting a pristine image \(\mathbf{x}\) with degradation from \(\mathbf{y}\).  Best to be viewed zoomed in.
    % In the first row, five distorted image patches
    % (\({\mathbf{y}}^{(j)}\)'s) containing different
    % image degradations are shown.  In the first column are their
    % corresponding pristine images \({\mathbf{x}}^{(i)}\)'s.
    % In each remaining grid is the synthetic distorted image generated by
    % transferring distortions from 
    % \({\mathbf{y}}^{(j)}\)
    % to 
    % \({\mathbf{x}}^{(i)}\). 
  }
  \label{fig:transfer}
\end{figure}

To test our model's generalizability to image semantics, we curated
a dataset of 300K Wikimedia Quality Images (WQIs), \hm{each of which is
inspected by Wikipedia or its sister projects' editors for high
technical and aesthetic quality}
\cite{wikicommon}.
% \footnote{This gathered dataset will be publicly released.} 
From this, we generated 40K training pairs using about 13K
images in the dataset.  Each pair consists of a pristine image
\(\mathbf{x}\) and a distorted image \(\mathbf{y}\), generated by
distorting  \(\mathbf{x}\) with distortions randomly selected from a
pool of degradation combinations designed according to
typical image processing pipelines.  
%We trained the model on 100 epochs
%using the Adam optimizer with an initial learning rate 1e-4 for \(\hat
%f\) and 1e-3 for other modules 
Details of the dataset and training
settings are in \supploc{4}.
\Cref{fig:transfer} showcases our model's ability to reproduce and transfer various degradations (e.g., Gaussian noise, blur, motion blur, JPEG artifacts) and their combinations across different images. 
Unlike \cite{bovik_2020}, our model generates diverse noise patterns, demonstrating both effective representation learning and output diversity.

\begin{figure}[t]
  \centering
  \includegraphics[width=0.98\columnwidth]{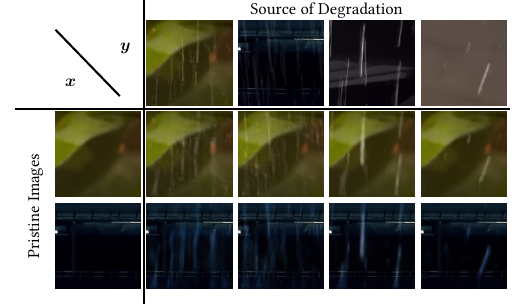}
  \caption{
    Realistic distortion transfer on the raindrop
    dataset \cite{rain_dataset}.
  }
  \label{fig:transfer_local}
\end{figure}
\begin{table}[t]%htbp]
\centering
% \sffamily
% \tiny % Make the font size smaller
\resizebox{\columnwidth}{!}{% Resize table to fit within the text width
	% \fontsize{10}{0}
	% \Huge
  \begin{tabular}{r|rrrrr}
\hline
& \textbf{MS-SSIM} $\uparrow$& \textbf{SSIM}$\uparrow$ & \textbf{LPIPS} $\downarrow$ & \textbf{DISTS} $\downarrow$ \\
\hline
Reproduce & 0.879 & 0.860 & 0.295 & 0.141 \\
Transfer & 0.875 & 0.856 & 0.306 & 0.147 \\
\hline
  \end{tabular}
}
\caption{
  Quantitative evaluation results of our model in reproducing image
  degradations.  
}
\label{tab:quan1}
\end{table}

By analyzing variances, we identified five active dimensions in
\(\mathbf{e}_g\) for the model trained on the distorted WQI dataset. 
\hm{%
  The model learns to disentangle degradation components
  and enables human to find interpretable meaning of each active latent
  dimension. 
}%

Authentic image degradations often include inhomogeneous elements that cannot be easily modeled by simple statistical models;
% Due to the challenges in collecting paired pristine-distorted datasets, such authentic datasets are scarce. 
for example, real-world raining effects \cite{rain_dataset}. As shown in \Cref{fig:transfer_local}, our model successfully reproduces and transfers these complex, authentic distortions.
Extra tests on the GoPro
dataset \cite{gopro} is in \supploc{3.2}.

Since this is the first universal image degradation model, it is
difficult to find comparable models.  Comprehensive ablation studies are available in \Cref{sec:abl}. 
To quantitatively evaluate our model's degradation reproduction and transfer capabilities, we created a test set of 2,000 image triplets $(\mathbf{x}^{(0)}, \mathbf{y}^{(0)}, \mathbf{y}^{(1)})$ from 4,000 pristine WQIs that were not
used for training. Here, \(\mathbf{x}^{(0)}\) is the pristine version of \(\mathbf{y}^{(0)}\), and \(\mathbf{y}^{(1)}\) is generated from a different pristine image using the same degradation as in \(\mathbf{y}^{(0)}\).
The reproduction score is calculated 
\modified{as 
  \(d \left( 
    \mathbf{y}^{(0)}
    ,
    \hat{f} \left( \mathbf{x}^{(0)}, \mathbf{e}_g^{(0)},
      \mathbf{e}_l^{(0)} \right)
    \right)
  \), 
  and the transfer score is calculated as
  \(d \left( 
    \mathbf{y}^{(0)}
    ,
    \hat{f} \left( \mathbf{x}^{(0)}, \mathbf{e}_g^{(1)},
      \mathbf{e}_l^{(1)} \right)
    \right)
  \), 
  where
  \(
  \mathbf{e}_{\cdot}^{(i)} = {e}_{\cdot} \left( \mathbf{y}^{(i)}
    \right)
  \).
}%
% by comparing 
% with
% \(\mathbf{x}^{(0)}\) distorted with
% degradation latents 
% \( (\mathbf{e}_g^{(0)}, \mathbf{e}_l^{(0)} ) \)
% obtained from \(\mathbf{y}^{(0)}\), and
% the transfer score is by comparing 
% \( (\mathbf{e}_g^{(0)}, \mathbf{e}_l^{(0)} ) \)
% We employ SSIM \cite{ssim}, MS-SSIM \cite{MS-SSIM}, LPIPS \cite{lpips}, and DISTS \cite{dists} as the metrics.
Results in \Cref{tab:quan1} demonstrate the model's high accuracy in degradation reproduction and transfer, indicating effective separation of degradation information from distorted images.

\subsection{Degradation latent disentanglement}

\begin{figure}[t]
  \centering
  \includegraphics[width=0.90\columnwidth]{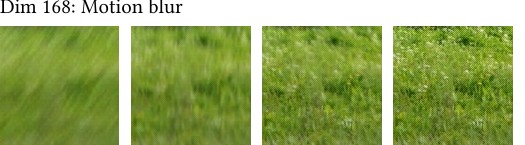}
  \caption{
    Demonstration of degradation latent disentangling by varying single
    entry in \(\mathbf{e}_g\) on a selected dimension. 
    %In each row, an active dimension of the degradation embedding is
    %varied by a fixed amount.  
    % The dimensions are ordered based on their standard
    % deviations.  
    %The degradations a single dimension controls is
    %labeled on the figure.  % \hl{fixed interval, linear space}
  }
  \label{fig:disentangle}
\end{figure}

% To demonstrate disentanglement, 
For demonstration,
we modified each active dimension of the degradation embedding for a pristine
test image. \Cref{fig:disentangle} shows visual results for one of the
active dimensions, with additional examples available in the
\supploc{3.3}.
Most degradation embedding dimensions control a group of degradations, aligning with our understanding of natural biases and
masking effects \cite{eonss} in the dataset. % \footnote{See \supplement for a detailed discussion.}. 

\subsection{Film grain encoding and transfer}

\begin{table}[t]%htbp]
\centering
% \sffamily
% \tiny % Make the font size smaller
\resizebox{\columnwidth}{!}{% Resize table to fit within the text width
\begin{tabular}{r|rrrrr}
\hline
& \textbf{MS-SSIM} $\uparrow$& \textbf{SSIM}$\uparrow$ & \textbf{LPIPS} $\downarrow$ & \textbf{DISTS} $\downarrow$ \\
\hline
Reproduce &  0.869 & 0.726 & 0.209 & 0.051  \\
Transfer &  0.872 & 0.728 & 0.209 & 0.053  \\
\hline
\end{tabular}
}
\caption{
  Quantitative evaluation results of our model in reproducing film grain.  
}
\label{tab:quan2}
\end{table}
\begin{figure}[t]
  \centering
  \includegraphics[width=0.98\columnwidth]{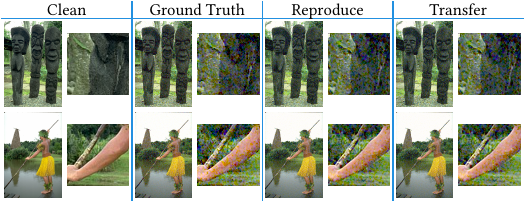}
  \caption{
    The result of film grain synthesis. 
  }
  \label{fig:grain}
\end{figure}

\begin{table}[t]%htbp]
\centering
\footnotesize

\begin{tabular}{r|rr|rr|rr}
\hline
\multicolumn{1}{c|}{} & \multicolumn{2}{c|}{\textbf{Accur. $\downarrow$}} & \multicolumn{2}{c|}{\textbf{Fidelity $\downarrow$}} & \multicolumn{2}{|c}{\textbf{Realism $\downarrow$}} \\ \hline
% \multicolumn{1}{c|}{} & \multirowcell{2}{w/o Ours\\w/ Ours} & \multirowcell{2}{w/o Ours\\w/ Ours} & \multirowcell{2}{w/o Ours\\w/ Ours} \\ \hline
\multicolumn{1}{c|}{} &\textbf{w/o} & \textbf{w/} & \textbf{w/o} & \textbf{w/} & \textbf{w/o} & \textbf{w/} \\
 &\textbf{Ours} & \textbf{Ours} & \textbf{Ours} & \textbf{Ours} & \textbf{Ours} & \textbf{Ours} \\ \hline
\multicolumn{7}{l}{\textbf{2 degradations}} \\ \hline
NA & .680 & \textbf{.513} & .424 & \textbf{.334} & 228.8 & \textbf{28.1} \\
AP & .506 & \textbf{.463} & .232 & \textbf{.219} & 45.1 & \textbf{16.9} \\
UA & .519 & \textbf{.512} & \textbf{.274} & .275 & 47.8 & \textbf{30.0} \\
NP & .713 & \textbf{.485} & .187 & \textbf{.081} & 221.9 & \textbf{20.6} \\
UN & .700 & \textbf{.564} & .273 & \textbf{.177} & 211.2 & \textbf{41.4} \\
UP & .541 & \textbf{.478} & .130 & \textbf{.097} & 57.7 & \textbf{38.8} \\
\hline
\multicolumn{7}{l}{\textbf{3 degradations}} \\ \hline
UNP & .715 & \textbf{.590} & .156 & \textbf{.082} & 215.6 & \textbf{60.9} \\
UAP & .588 & \textbf{.526} & .169 & \textbf{.115} & 80.5 & \textbf{34.8} \\
UNA & .669 & \textbf{.569} & .407 & \textbf{.298} & 175.5 & \textbf{34.5} \\
NAP & .686 & \textbf{.508} & .282 & \textbf{.182} & 213.7 & \textbf{25.3} \\
\hline
\multicolumn{7}{l}{\textbf{4 degradations}} \\ \hline
UNAP & .666 & \textbf{.560} & .251 & \textbf{.141} & 147.4 & \textbf{31.7} \\
\hline
\end{tabular}

\caption{
  Accuracy (LPIPS \cite{lpips}), fidelity (LPIPS \cite{lpips}) and
  realism (pFID \cite{pfid}) comparison of the same image restoration method \cite{robust_stylegan} 
  under blind settings w/o  and w/ our universal degradation model on the restoration of images degraded by complex combinations of distortions. 
  The method that has the best score in each category is shown in
  \textbf{bold fonts}.
}
\label{tab:comparison_single}
\end{table}

\modified{It is a common practice in the film-making industry to simulate film grains for digital movies, as directors and producers consider it essential 
for creating cinematic atmosphere and immersive viewing experiences \cite{av1_filmgrain}.}
%It is essential to accurately reproduce and synthesize film grains for
%the film industry, as it can enhance the ``cinematic'' feeling of a movie, which
%creates a more immersive viewing experience
%%%%%%%%%%%%%%%%%
We fine-tune our model on the FilmGrainStyle \cite{filmgrain_style}
dataset and evaluate it using their test set, both containing
professionally inspected grainy images. 
The film industry has two primary requirements for film grain synthesis
(a) encoding and reproducing film grains from grainy images to grain-free versions, 
and 
(b) to transfer realistic film grain
from existing images to new ones. 
\Cref{tab:quan2} shows quantitative results of both tasks with visuals
in \Cref{fig:grain}, from which realistic film grain
patterns can be observed. 
To evaluate transfer performance, we simulate real application
scenarios by transferring grains from a training image (with
similar grain style as indicated by the
test set's label) to a grain-free test image, and compare it with the
ground-truth grainy image.
The comparable performance between transfer and reproduction scores demonstrates 
our model's effectiveness in disentangling degradation information.
Notably, despite being designed as a universal model, our approach outperforms 
the specialized baseline model from \cite{filmgrain_style} in reproduction scores.

% We then use the supplied test set \cite{filmgrain_style} to evaluate our model's
% encoding-decoding performance, as well as the performance of
% synthesizing grains on new contents by transferring existing grains.  
% To evaluate the transfer performance, we simulate real application
% scenarios by transfer grains from an image
% in the training set (which has similar grain style as indicated by the
% test set's label) to a clean test image, and compare it with the
% ground-truth grainy image.  Quantitative results are shown in
% \Cref{tab:quan2} with visuals  in 

\subsection{Inversion-based image restoration}

\modified{
A major limitation of existing Inversion-Based Image Restoration (IBIR)
methods is their reliance on an impractical \textit{non-blind}
assumption: degradation parameters must be known during inference
\cite{lbrgm,robust_stylegan,dps}, an assumption rarely aligns with
real-world scenarios. To address this, our model can be plugged into
existing non-blind IBIR methods, converting them into blind restoration
approaches. We upgraded two state-of-the-art non-blind restoration
methods with our model: Robust StyleGAN Inversion (RSG)
\cite{robust_stylegan} and Diffusion Posterior Sampling (DPS)
\cite{dps}. In this section, we focus on RSG due to its simplicity and
lower computational cost. While our computational resources limit the
scale of experiments for DPS, qualitative results are provided in the
\supploc{5.4}.
}

%\modified{A significant limitation of existing Inversion-Based Image Restoration (IBIR) methods is their impractical \textit{non-blind} assumption: degradation parameters must be known during inference \cite{lbrgm,robust_stylegan,dps}, which rarely occurs in real-world scenarios. To mitigate this, our model can be plugged into existing non-blind IBIR methods to convert them to blind ones. We have augmented two state-of-the-art non-blind restoration methods with our model: Robust StyleGAN Inversion (RSG) \cite{robust_stylegan} and Diffusion Posterior Sampling (DPS) \cite{dps}. The former is used for presentation in this section due to its simplicity and lower computational cost. Despite our computational resources limit the scale of the experiments for DPS, qualitative results for DPS are available in the \supplement.}

A \modified{generative model} \(G: \mathbf{w} \to \mathbf{x}\)
maps a randomly sampled input \(\mathbf{w}\) from a known distribution
\(\mathcal{W}\) to a generated image \(\mathbf{x}\).
%\footnote{For the
%simplicity of notations, ``\(\mathcal{W}\)'' denotes both the space and
%distribution of variables.}  
\modified{%
For IBIR tasks, the output high-quality image \(G\left(
\mathbf{w} \right)\) must be re-distorted for comparison.  
}%
In particular, 
\citet{robust_stylegan} propose an optimization method to empirically
solve 
\begin{equation}
  \min\limits_{\mathbf{w}}\text{ LPIPS}\left( \mathbf{y},D_{\theta_{d}}\left( G\left( \mathbf{w} \right) \right) \right),
\end{equation}
where 
\(D_{\theta_{d}}\) is the continuous approximation of the
degradation operation parametered by the already recorded \(\theta_{d}\).
To make it a \textit{blind} method, we replace \(D_{\theta_{d}}\)
with our model:
\begin{equation}
  \min\limits_{\mathbf{w},\mathbf{e}_g,\mathbf{e}_l}
  \mathbb{E}_{\mathbf{n}}
  \text{LPIPS}
  \left( \mathbf{y},\hat{f}\left(G\left( \mathbf{w}
  \right),\mathbf{e}_g, \textbf{e}_l,\mathbf{n}\right) \right),
\end{equation}
where \(\hat{f}\) is our degradation synthesis module, and
\(\mathbf{e}_g\) and \(\mathbf{e}_l\) are predicted by our model
from \(\mathbf{y}\).

\begin{figure}[t]
  \centering
  \includegraphics[width=0.99\columnwidth]{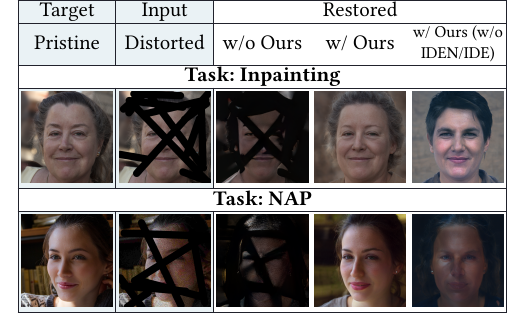}
  \caption{Image restoration results for a single/complex combination of
    homo- and inhomogeneous degradations. The naive blind GAN inversion
    method, without universal degradation modeling, produces unsatisfactory results. In contrast, with our universal degradation model, the otherwise identical model achieves significantly improved results, while remaining fully blind. 
  }
  \label{fig:restoration}
\end{figure}

% We augment RSG \cite{robust_stylegan} with our
% degradation model to convert it 
% to a blind image restoration method. 
We benchmark the augmented method on the categories and the test set (FFHQ-X)
%\footnote{Since all
%images in publicly available FFHQ dataset are used for training the StyleGAN model
%\cite{stylegan2}, a separate dataset FFHQ-X with properly aligned images
%is needed for testing.}
proposed in \cite{robust_stylegan}. 
\modified{%
  The test set consists of face images, 
  due to the underlying GAN model's limitations. 
  However, our approach can be applied to any generative model and image content
  (results using DPS \cite{dps} on ImageNet \cite{imagenet} are available in the
  \supploc{5.4}).
}%
For fair comparison, we only show performance against the only
blind GAN inversion-based IR method (\textit{i.e.}, without
re-degradation). 
%against other GAN inversion-based approaches: \cite{robust_stylegan}
%(RSG), PULSE (PUL) \cite{pulse}, and L-BRGM (LBR) \cite{lbrgm}
The benchmarked categories include single degradations (\textbf{U}psampling, de\textbf{N}oising, de\textbf{A}rtifacting, and in\textbf{P}ainting) and their combinations.
% All benchmarked methods are designed to invert the same StyleGAN model
% \cite{stylegan2}, which use FFHQ \cite{stylegan} as the training set. 
% Our degradation model is trained on the FFHQ dataset distorted with benchmarked
% degradations of random parameters.
\Cref{fig:restoration} \hm{shows examples of restoration results from our
model} \hm{and the naive blind restoration
method}. 
\hm{Without utilizing our model for redegradation, the result produced by
  blind GAN-inversion method is visually unacceptable.
}%
The quantitative results are shown in \cref{tab:comparison_single}. Due to space
limitations, we only present results for multi-degradation restoration; the full table is available in the
\supploc{5.3}.
% \hm{Although converting a non-blind restoration method to a blind one is
%   a trade-off between image quality and application flexibility, 
% the performance drop of our model from} \cite{robust_stylegan} 
% \hm{is rather small.}
% Our model performs competitively with state-of-the-art non-blind GAN
% inversion-based image restoration models. More importantly, 
When
compared under the same conditions, our model outperforms the vanilla
blind restoration method in accuracy, fidelity, and
in realism for most tasks.

\notebox{
  Check the last two sentences
}

\subsection{Ablation Studies}
\label{sec:abl}

\begin{figure}[t]
  \centering
  \includegraphics[width=0.99\columnwidth]{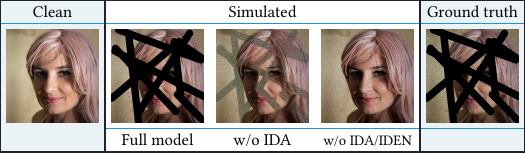}
  \caption{
    Examples of degradation transfer from mixed sources.
    Without the IDA layer or IDEN, the model is unable to properly
    reproduce inhomogeneous degradation.  
    Inputs to the models (\(\mathbf{x}, \mathbf{y}_1, \mathbf{y}_2\)) are shown in \cref{fig:abl_2}.
    % using our full model and ablated models.  The 
  }
  \label{fig:abl_2r}
\end{figure}

We construct a test dataset from FFHQ-X \cite{robust_stylegan} using their proposed degradations to verify our newly introduced modules. For each pristine image \(\mathbf{x}^{(0)}\), we compute the following transfer scores:
\begin{itemize}
  \item Direct: Transferring both homogeneous and inhomogeneous
    degradation  from a distorted image
    \(\mathbf{y}^{(1)}\) with different contents. 
  \item Global-only: Transferring both homogeneous 
    degradation  from a distorted image 
    \(\mathbf{y}^{(2)}\) with different content
    that only contain
    homogeneous degradation
  \item Mixed: Transferring homogeneous 
    degradation from one distorted image 
    \(\mathbf{y}^{(1)}\) 
    and 
    inhomogeneous degradation from 
    another distorted image 
    \(\mathbf{y}^{(3)}\).  All three involved images have different
    contents. 
\end{itemize}

Since all three cases use distortions synthesized from Python scripts, we are
able to generate ground-truth distorted images to evaluate our performance. 
The distorted FFHQ dataset constructed in the previous subsection is
used to train the models analyzed in the ablation study.

\textbf{Disentangling-by-compression }
To benchmark the disentangle-by-compression method, we removed the
entropy regularization loss functions. The results, shown in
\Cref{tab:abl_1}, indicate that this change led to decreased performance in degradation transfer, with a more
pronounced effect on mixed transfer. This demonstrates the effectiveness
of the disentangle-by-compression method in separating degradation
information from distorted images and disentangling local from global
degradations.

\textbf{IDA layer and IDEN}
We introduced the IDE network and IDA layer for modeling inhomogeneous
degradations. Our ablation study incrementally removed these components
from the full model. The results, presented in \Cref{tab:abl_2} with a
visual example in \Cref{fig:abl_2r}, show that the IDA layer and IDE
network not only enhance the reproduction and transfer of inhomogeneous
degradations but also improve homogeneous degradation reproduction.
Additional visual examples are available in \supploc{3.4}.

\begin{table}[t]%htbp]
\centering
% \sffamily
% \tiny % Make the font size smaller
% \resizebox{\columnwidth}{!}{% Resize table to fit within the text width
	% \fontsize{10}{0}
	% \Huge
\footnotesize
\begin{tabular}{r|llll}
\hline
\textbf{Model} & \textbf{Direct} & \textbf{Mixed} \\
\hline
% benchmark_res/full/scores.json
Full & .271 & .286 \\
% benchmark_res/abl1/scores.json
No disentanglement (w/o entropy loss) & .290 (+.018) & .334 (+.048) \\
\hline
\end{tabular}
% }
\caption{
  Ablation study results on the disentangle effects of the
  disentangle-by-compression loss on degradation transfer.
  LPIPS scores ($\downarrow$) for each modified model are shown. 
}
\label{tab:abl_1}
\end{table}

\begin{table}[t]%htbp]
  % Let it flow, which may save some space
\centering
% \sffamily
% \tiny % Make the font size smaller
% \footnotesize
% \tiny
% \small
\resizebox{\columnwidth}{!}{% Resize table to fit within the text width
	% \fontsize{10}{0}
	% \Huge
\begin{tabular}{r|llll}
\hline
\textbf{Model} & \textbf{Global-only} & \textbf{Direct} &
\textbf{Mixed} \\
\hline
% benchmark_res/full/scores.json
Full & .388 & .271 & .286 \\
% benchmark_res/abl2/scores.json
% No IDA & .389 (+.001) & .275 (+.004) & .288 (+.002) \\
% benchmark_res/abl2a/scores.json
No IDA & .404 (+.016) & .298 (+.027) & .356 (+.070) \\
% benchmark_res/abl3/scores.json
No IDA, No IDEN & .394 (+.006) & .577 (+.306) & .588 (+.302) \\
\hline
\end{tabular}
}
\caption{
  Ablation study results on the effects of inhomogeneous
  degradation-related structures on degradation transfer.
  LPIPS scores ($\downarrow$) for each modified model are shown. 
}
\label{tab:abl_2}
\end{table}

\section{Conclusion}

We present the first universal image degradation model. The proposed disentangling-by-compression strategy enables the model to effectively separate distortion information from the distorted
image's content while encouraging independence among the components of
distortion embedding. The introduction of the IDA-SFT layer and the IDEN
extends our model's capability to handle both homogeneous and
inhomogeneous degradations, significantly broadening its applicability.
Notably, our model transforms the non-blind restoration methods
\cite{robust_stylegan,dps} into blind ones, achieving competitive performance even without any distortion information.

\section*{Acknowledgements}

This work is supported in part by National Sciences and Engineering
Research Council of Canada under the Canada Research Chair and Discovery
Grant programs.

{
    \small
    \bibliographystyle{ieeenat_fullname}
    \bibliography{main}
}

\end{document}